%% file: iclr2027_conference.tex
\documentclass{article} 
\usepackage{iclr2027_conference,times}
\usepackage{algorithm}
\usepackage{algpseudocode}
\usepackage{graphicx}
\usepackage{booktabs}
\usepackage{amssymb} 
\usepackage{multirow}
\usepackage{placeins}
\usepackage{wrapfig}

\input{math_commands.tex}

\usepackage{hyperref}
\usepackage{url}
\usepackage{xspace}

\newcommand{\seed}{x_0}

\DeclareMathOperator{\simop}{sim}
\title{\modelname{}: Inference-Time Refinement and Self-Distillation\\ for Controllable Molecular Optimization}

\author{
Jia Yi Zhang$^{1}$\thanks{Equal contribution} ,
Shivanshu Shekhar$^{1}$\footnotemark[1] ,
Allison McLevige$^{1}$,
Tong Zhang$^{1}$\\
$^{1}$University of Illinois Urbana-Champaign \\}

\newcommand{\modelname}{ReMedI\xspace}
\iclrfinalcopy 
\begin{document}

\maketitle
\lhead{Preprint}
\begin{abstract}

In drug development, lead optimization must improve competing properties, such as increasing permeability and reducing toxicity, while limiting scaffold changes that could compromise the lead’s potency. RL-based post-training, such as C-MORAL, addresses this problem by fine-tuning molecular language models to directly optimize molecular leads for objectives of interest, which relies on expensive task-specific datasets and faces challenges in achieving multi-objective optimization (MOO). We argue a pretrained model may already contain useful molecular edits that zero-shot generation cannot efficiently reach. Motivated by the limited search capabilities of zero-shot generation, we develop an efficient Metropolis--Hastings search strategy to better explore the proposal distribution. We propose \modelname{}, which uses the molecular language model as a base proposer for inference-time alignment. Given a lead and a set of multi-objective tasks, the model proposes edits conditioned on the current molecule, while Metropolis--Hastings refinement guides sequential edits under an objective for property improvement, preservation, and lead similarity. On both C-MuMOInstruct and SMDD benchmarks, refinement improves frozen base models across three backbones, reaching significantly higher multi-objective optimization success rates compared to baselines. These search traces can then distilled into the base model which improves zero-shot generation and subsequent search using $\approx7.8\times$ less training molecular leads than that of C-MORAL. \modelname{} achieves $\approx2.8\times$ the strict-success coverage of RL baseline C-MORAL, including on out-of-distribution tasks, with matched lead-relative similarity.
\end{abstract}

\section{Introduction}

Chemical language models (CLMs) have increasingly become popular for text-guided molecular generation~\citep{edwards2022translationmoleculesnaturallanguage}, multi-property optimization~\citep{gellmo,cmumo}, and the design of functional, makeable molecules~\citep{mclm}. A central application of CLMs is lead optimization, which starts with a promising molecule (lead) and seeks local edits that improve deficient properties, such as permeability or toxicity, while preserving satisfactory properties and maintaining lead-relative similarity. However, current CLMs do not reliably produce molecules that meet all property targets without drifting off-scaffold. RL post-training and reference-guided policy optimization methods try to align CLMs to property rewards to satisfy MOO ~\citep{cmoral,li2026referenceguidedpolicyoptimizationmolecular}. For instance, C-MORAL combines group-based policy optimization with property-specific score transformations and nonlinear reward aggregation to improve selective molecular editing~\citep{cmoral}. This approach relies on task-specific RL adapters and editing data, which are costly to develop and can limit transfer to new combinations of objectives. Since chemists often optimize several properties together, the challenge therefore is to meet every property requirement and maintaining acceptable lead-relative similarity without adapting a separate model for each task.

Inference-time alignment offers an alternative by guiding generation while holding the model weights fixed. Existing approaches include similarity-guided decoding~\citep{navratil2025gpmolformersimtesttimemolecular} and iterative editing~\citep{mclm}. For instance, GP-MoLFormer-Sim guides decoding through reference similarity and uses an outer genetic algorithm for property optimization~\citep{navratil2025gpmolformersimtesttimemolecular}, but its decoding guidance does not directly encode which properties should improve or remain unchanged. MARS demonstrates that multi-property molecular optimization can benefit from iterative search over intermediate molecular states~\citep{mars}. We build on this principle in \modelname{} by using an instruction-conditioned language model to propose successive edits with a search algorithm which determines which states to retain. 

We introduce \modelname{}, which uses a pretrained molecular language model to propose local edits and Metropolis--Hastings (MH) refinement to guide the search. Its lead-relative objective balances property improvement, preservation, and similarity to the lead. Because the proposal remains frozen during refinement, the improvement--similarity trade-off can be adjusted at inference time without task-specific RL adapters. Thus our method allows chemists to explore pharmacological spaces that are hard to reach via zero-shot prompting. On C-MuMOInstruct and SMDD, refinement improves success across GeLLMO-Llama, GeLLMO-Mistral, and mCLM-4B, and \modelname{} achieves higher strict-success coverage than C-MORAL on matched GeLLMO-Mistral leads, including on combinations of properties held out during training. 

We show that in \modelname{} a single shared proposal adapter trained from only 500 lead--task instances achieves performance comparable to C-MORAL, which relies on five task-specific policies trained from 3,883 lead--task prompts. Our traces contain successful endpoints and intermediate states, so training does not require a separate specialized editing dataset. Search improves candidates at inference time and supplies examples that strengthen later generation and refinement.

\begin{figure}[t]
\centering
\includegraphics[width=\columnwidth]{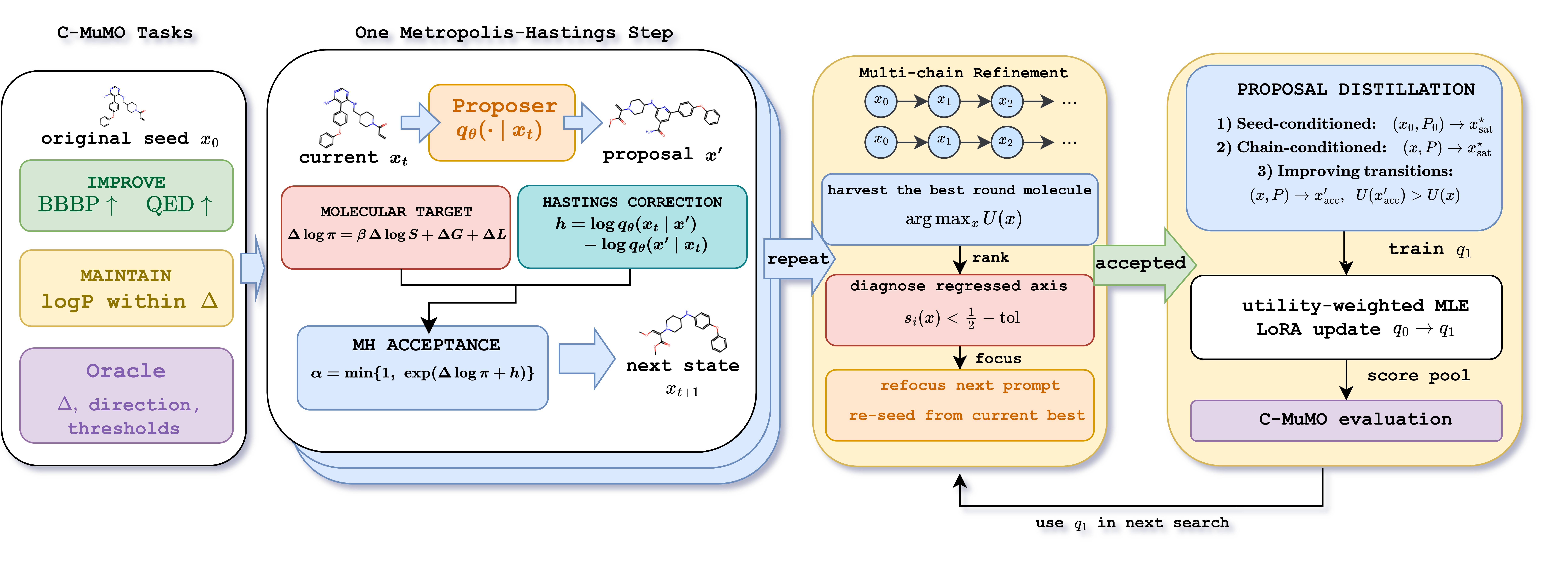}
\caption{ReMedI separates the molecular edit proposal from the inference-time objective. Refinement discovers successful trajectories; lead- and state-conditioned examples from those trajectories train a shared proposal for later generation and search.}
\label{fig:workflow}
\end{figure}

\paragraph{Contributions.}
\begin{itemize}
    \item \textbf{Inference-time molecular alignment.}
    We use a chemical language model (CLM) to propose edits. Metropolis--Hastings (MH) accepts or rejects each proposal using a lead-relative objective for property improvement, preservation, and similarity, together with the corresponding LM probabilities. These priorities can be changed at inference time without retraining the model.
    \item \textbf{Learning from refinement.}
    We convert search-discovered endpoints, intermediate states, and improving transitions into supervised training pairs and distill them back into the base CLM which improves zero-shot generation and subsequent MH refinements.
    \item \textbf{Success under similarity requirements.}
    \modelname{} achieves $\approx2.8\times$ the strict-success coverage of C-MORAL on matched  GeLLMO-Mistral leads. On C-MuMOInstruct, MH increases frozen-model SOR by 66--143\% relative to zero-shot generation across three backbones and IND/OOD splits; on SMDD, it improves SOR by 12--50\%. 
\end{itemize}

\section{Related Works}
\textbf{Post-training for molecular optimization.}
Molecular LLMs support text-guided generation, editing, and multi-property optimization~\citep{acl_survey}. Early molecule--text models such as MolT5 and MoleculeSTM connect molecular strings to natural language~\citep{edwards2022translationmoleculesnaturallanguage,liu2023multimodal}, while instruction- and interaction-based systems such as LlaSMol, PEIT-LLM, ChatDrug, and DrugAssist improve chemistry task following, property-conditioned generation, or conversational molecular editing~\citep{yu2024llasmoladvancinglargelanguage,lin2026propertyenhancedinstructiontuning,liu2023chatgptpoweredconversationaldrugediting,ye2023drugassistlargelanguagemodel}. For controllable multi-property optimization, Prompt-MolOpt steers property-specific transformations with prompt-conditioned embeddings~\citep{wu2024promptmolopt}, while GeLLMO-C and C-MuMOInstruct train models to follow multi-property instructions and prevent property regressions~\citep{gellmo,cmumo}. Earlier molecular optimization methods also applied reward-guided reinforcement learning to sequence generation and direct molecular edits~\citep{olivecrona2017molecular,popova2018deepreinforcement,zhou2019moldqn}. More broadly, post-training methods modify the generative policy through supervised fine-tuning, preference optimization, or reinforcement learning~\citep{ouyang2022instructgpt,rafailov2023dpo,schulman2017ppo}. C-MORAL uses GRPO/GDPO-style optimization to align molecular LLM outputs with multi-objective constraints~\citep{cmoral,shao2024deepseekmath}, while RePO uses reference-guided policy optimization~\citep{li2026referenceguidedpolicyoptimizationmolecular}. These approaches adapt the model through specialized training. \modelname{} instead keeps the CLM fixed during refinement and specifies the improvement--similarity trade-off through an explicit objective at inference time.

\textbf{Inference-time alignment.}
A smaller line of work shifts optimization to inference time. MARS uses Markov-chain search for multi-objective molecular optimization with graph neural networks~\citep{mars}, while GP-MoLFormer-Sim provides training-free similarity-guided decoding for chemical language models~\citep{navratil2025gpmolformersimtesttimemolecular}. \modelname{} uses the CLM as a state-dependent proposal within a Metropolis--Hastings refinement loop~\citep{metropolis1953,hastings1970}, with an explicit target that favors requested property improvements, preservation of guarded properties, and similarity to the lead. Search can also improve the proposal itself. Expert Iteration formalizes the search-and-distill pattern~\citep{anthony2017thinkingfastslowdeep}, and we train a shared proposal adapter on successful refinement endpoints and intermediate-state examples, reconditioning multi-step solutions on the original lead as one-step targets. This combines inference-time refinement with proposal improvement and tests whether search-derived supervision transfers to unseen objective combinations.

\section{Preliminary}
\label{sec:prelim}
\paragraph{Metropolis--Hastings.}
Metropolis--Hastings (MH) draws samples from a distribution known only up to a normalizing constant. Given an unnormalized target $\pi$ and a proposal $q(\cdot\mid x)$, from the current state $x$ it draws a candidate $y\sim q(\cdot\mid x)$ and accepts it with probability 
\begin{equation}
A(x\rightarrow y)=\min\!\left(1,\; \frac{\pi(y)}{\pi(x)}\,\frac{q(x\mid y)}{q(y\mid x)}\right),
\label{eq:mh}
\end{equation}
where the second factor, the Hastings correction, offsets the asymmetry of a state-dependent proposal (a forward edit likely from $x$ need not be equally likely in reverse). Since $\pi$ appears only through a ratio, its normalizing constant cancels. The proposal controls exploration, while $\pi$ determines the target distribution. The proposal can be changed or learned without changing that target, provided the Hastings correction is retained and the resulting chain remains irreducible and aperiodic over the target's support.

\paragraph{Nash bargaining and Pareto optimality.}
Let $\mathcal X$ be the feasible outcome set. Consider $m$ objectives with utilities $u_i(x)$ and a reference (disagreement) point $d\in\mathbb{R}^m$ that specifies the baseline value for each objective. An outcome $x$ is Pareto-optimal if there is no feasible $x'$ such that $u_i(x')\ge u_i(x)$ for every $i$, with strict inequality for at least one objective. In other words, no objective can be improved without sacrificing another. If the feasible utility set is compact and convex and contains a point that strictly improves every objective over $d$, the Nash bargaining solution \cite{nash} has a unique utility vector, attained by any maximizer $x^{\mathrm{NBS}}$ of
\begin{equation}
x^{\mathrm{NBS}}\in\argmax_{\substack{x\in\mathcal X\\ u_i(x)\ge d_i\ \forall i}}\ \prod_{i=1}^{m}\big(u_i(x)-d_i\big),
\end{equation}
which maximizes the \emph{product} of gains over the reference point rather than their weighted sum. Equivalently, it maximizes $\sum_i\log\big(u_i(x)-d_i\big)$. This makes the solution invariant to positive affine rescaling of an objective when its disagreement value is transformed in the same way, and favors balanced gains across objectives. Any outcome that leaves one objective at its reference value has a zero product, regardless of gains in the other objectives. 
\section{Method}
\label{sec:method}
\begin{algorithm}[t]
\caption{Focused refinement and trace generation. The inner chain is given in Alg.~\ref{alg:mhchain}.}
\label{alg:refine}
\begin{algorithmic}[1]
\Require lead $\seed$, task requirements $t$, base prompt $P_0(t)$, valid proposal $q_\theta^{\mathrm{valid}}$, stages $R$, chains $C$, steps $N$, similarity target $\tau_s$, weights $\beta_{S,0},\beta_G,\beta_L$, annealing rate $\rho$
\Ensure incumbent $x^\star$; transition log $\mathcal T$
\State $x^\star\gets\seed$;\; $\phi\gets$ similarity;\; $P\gets P_{\phi}(t)$;\; $\textsc{Tried}\gets\varnothing$;\; $\mathcal T\gets\varnothing$
\For{$j=1,\dots,R$}
\State $\mathcal V\gets\{x^\star\}$
\For{$c=1,\dots,C$}
\State $\mathcal V_c,\mathcal T_c\gets\Call{MHChain}{x^\star,\seed,P,N,q_\theta^{\mathrm{valid}},\beta_{S,0},\rho,\beta_G,\beta_L,\tau_s}$
\State $\mathcal V\gets\mathcal V\cup\mathcal V_c$;\; $\mathcal T\gets\mathcal T\cup\mathcal T_c$
\EndFor
\State $\hat{x}\gets$ best-ranked candidate in $\mathcal V$ under the criteria above
\If{$\hat{x}$ is preferred to $x^\star$ under the criteria above} $x^\star\gets\hat{x}$;\; $\textsc{Tried}\gets\varnothing$
\ElsIf{$\phi$ is a property index} $\textsc{Tried}\gets\textsc{Tried}\cup\{\phi\}$
\EndIf
\If{$\mathcal R_t(x^\star)=\varnothing$ \textbf{and} $\simop(x^\star,\seed)\ge\tau_s$} \textbf{break} \EndIf
\State $\mathcal F\gets\mathcal R_t(x^\star)\setminus\textsc{Tried}$
\If{$\mathcal F\neq\varnothing$} $\phi\gets\argmin_{i\in\mathcal F}s_i(x^\star)$ \Comment{worst untried improve-property}
\Else\ $\phi\gets$ similarity \EndIf
\State $P\gets P_{\phi}(t)$
\EndFor
\State \Return $x^\star,\mathcal T$
\end{algorithmic}
\end{algorithm}

\paragraph{Scoring and target.}
Let $q_{\theta}$ denote the LM's conditional proposal distribution, $\seed$ the lead, $t$ the task requirements, and $P_0(t)$ the original prompt specifying those requirements. Let $p(x)\in\mathbb{R}^m$ be the oracle's property values and $d=p(\seed)$ the lead's values. The task specifies which properties to improve and which to guard. We denote these sets by $\mathcal I$ and $\mathcal G$, and use $\mathrm{sgn}_i\in\{+1,-1\}$ for the requested direction of property $i$. We map each requested property to a lead-centred satisfaction
\begin{equation}
s_i(x)=\operatorname{clip}_{[\epsilon,1-\epsilon]}\!\left(\sigma\!\big(k_i\,\mathrm{sgn}_i\,(p_i(x)-d_i)\big)\right),
\qquad
\sigma(z)=\tfrac{1}{1+e^{-z}},
\label{eq:sat}
\end{equation}
where $k_i>0$ controls the response scale for property $i$, and $0<\epsilon<1/2$ keeps each satisfaction away from zero and one so its logarithm remains bounded. We aggregate the satisfactions of properties to improve as $S(x)=\prod_{i\in\mathcal I}s_i(x)$, so gains on one property cannot fully mask poor satisfaction on another. For guarded properties, the penalty $G(x)$ activates only when a property declines beyond its allowed margin. The similarity penalty $L(x)$ activates when similarity to the lead falls below $\tau_s$.

\begin{equation}
G(x)=-\!\sum_{i\in\mathcal G}\max\!\big(0,-\tau_i^g-\mathrm{sgn}_i(p_i(x)-d_i)\big),
\qquad
L(x)=-\max\!\big(0,\tau_s-\simop(x,\seed)\big).
\label{eq:pen}
\end{equation}

Here $\tau_i^g$ is the allowed decline for guarded property $i$, and $\tau_s$ is the minimum lead similarity. Combining the three terms gives the utility score and target distribution,
\begin{equation}
\begin{aligned}
U(x)&=\beta_S\log S(x)+\beta_G\,G(x)+\beta_L\,L(x),\\
\pi(x)\propto{}&\exp\!\big(U(x)\big),
\end{aligned}
\label{eq:target}
\end{equation}
Here $\log S(x)=\sum_{i\in\mathcal I}\log s_i(x)$ is the sum of per-property log-satisfactions and the log of the Nash product. The positive weights $\beta_S$, $\beta_G$, and $\beta_L$ control improvement, guarding, and similarity, respectively.

\paragraph{Sampling and focused refinement.}
At each refinement stage, we set the focused prompt to $P=P_{\phi}(t)=P_0(t)\oplus\mathrm{clause}(\phi)$, where $\phi$ identifies the property or similarity constraint to focus on. The model proposes edits from $q_{\theta}^{\mathrm{valid}}(\cdot\mid x,P)$. For a proposed move from $x$ to $x'$, write $\Delta F=F(x')-F(x)$ for any score $F$. At MH iteration $n$, we anneal the satisfaction weight as $\beta_{S,n}=\beta_{S,0}\rho^{n-1}$, giving the step-specific target $\pi_n(x)\propto\exp\!\big(\beta_{S,n}\log S(x)+\beta_G G(x)+\beta_L L(x)\big)$. The Metropolis--Hastings acceptance ratio is
\begin{equation}
\log r=
\underbrace{\beta_{S,n}\,\Delta\log S+\beta_G\,\Delta G+\beta_L\,\Delta L}_{\Delta\log\pi_n}
+
\underbrace{\log q_{\theta}^{\mathrm{valid}}(x\mid x',P)-\log q_{\theta}^{\mathrm{valid}}(x'\mid x,P)}_{\log h},
\label{eq:accept}
\end{equation}
We accept the move with probability $\alpha=\min\{1,e^{\log r}\}$. Each iteration applies the MH rule for its step-specific target $\pi_n$. We trim and parse proposals to valid molecules and retry invalid draws. The focus prompt guides which edits the model proposes, while the target score is determined by $S$, $G$, and $L$. The acceptance ratio accounts for the prompt by comparing the probabilities of proposing $x'$ from $x$ and $x$ from $x'$. Computing the target ratio requires oracle scores for both states.

\paragraph{Proposal distillation.}
The search produces traces of proposed edits and their outcomes. We use these traces as supervision to train the proposal model for future searches, curating a corpus $\mathcal D$ of context--molecule pairs with three types of examples. \emph{Seed-conditioned} examples pair the lead and its task prompt $(\seed,P_0(t))$ with a satisfying endpoint, teaching the model to generate directly from the lead. \emph{Chain-conditioned} examples pair an intermediate state and its focused task prompt $(x,P_{\phi}(t))$ with a satisfying endpoint reached from that state. We also include some \emph{accepted transitions} that do not meet every requirement, so the model learns incremental edits that keep refinement moving. We fit $q_\theta$ by weighted maximum likelihood,
\begin{equation}
\mathcal L(\theta)=-\!\!\sum_{(c,x')\in\mathcal D} w(x')\,\log q_\theta(x'\mid c),
\qquad
w(x')\propto\exp\!\big(U(x')/\beta_{\mathrm{dist}}\big),
\label{eq:distill}
\end{equation}

where the positive temperature $\beta_{\mathrm{dist}}$ controls how strongly the weights favor higher-scoring targets. Seed- and chain-conditioned examples match the two proposal contexts used at inference, so one distilled model can improve both direct generation and refinement. The process can be repeated by generating new traces with the current proposal and refitting Eq.~\ref{eq:distill}.

Let $\mathcal R_t(x)$ be the unmet improvement requirements for task $t$. We prefer candidates with fewer unmet requirements, then those meeting $\tau_s$, then higher utility $U(x)$. The focus $\phi$ is the least-satisfied unmet requirement, or similarity if none remain, and its clause tells the model what to focus on.


\begin{algorithm}[t]
\caption{\textsc{MHChain} performs one refinement chain of length $N$.}
\label{alg:mhchain}
\begin{algorithmic}[1]
\Function{MHChain}{$x_0,P,N,q_\theta^{\mathrm{valid}},\beta_{S,0},\rho,\beta_G,\beta_L,\tau_s$}
\State $x\gets x_0$;\; $\mathcal V\gets\{x\}$;\; $\mathcal T\gets\varnothing$
\For{$n=1,\dots,N$}
\State $\beta_{S,n}\gets\beta_{S,0}\rho^{n-1}$
\State $x'\sim q_\theta^{\mathrm{valid}}(\cdot\mid x,P)$
\State $\Delta\log\pi_n\gets\beta_{S,n}\,\Delta\log S+\beta_G\,\Delta G+\beta_L\,\Delta L$;\quad $\log h\gets\log q_\theta^{\mathrm{valid}}(x\mid x',P)-\log q_\theta^{\mathrm{valid}}(x'\mid x,P)$
\State $\log r\gets\Delta\log\pi_n+\log h$;\quad $\alpha\gets\min\{1,\exp(\log r)\}$
\State $x_{\mathrm{prev}}\gets x$;\quad $b\sim\operatorname{Bernoulli}(\alpha)$
\If{$b=1$} $x\gets x'$;\; $\mathcal V\gets\mathcal V\cup\{x\}$ \EndIf
\State $\mathcal T\gets\mathcal T\cup\{(\seed,x_{\mathrm{prev}},P,x',U(x'),\log q_\theta^{\mathrm{valid}}(x'\mid x_{\mathrm{prev}},P),\log h,\alpha,b)\}$
\EndFor
\State \Return $\mathcal V,\mathcal T$
\EndFunction
\end{algorithmic}
\end{algorithm}

\section{Experiments}
\label{sec:exp}

\subsection{Setup}

\paragraph{Models and benchmarks.}
We evaluate \modelname{} primarily on C-MuMOInstruct, which contains ten multi-property editing tasks: five training-seen property combinations (IND) and five held-out combinations (OOD)~\citep{cmumo}. We use GeLLMO-C Llama-3.1-8B, GeLLMO-C Mistral-7B, and mCLM-4B as proposal models~\citep{gellmo,mclm}, with property evaluation provided by ADMET-AI and PyTDC~\citep{swanson2024admetai,pytdc}. IND and OOD follow the benchmark's original task split; our shared proposal adapter is distilled only from trajectories spanning the five IND tasks. We additionally evaluate \modelname{} on 314 instances from SMDD-Bench Task 4 (Lead Optimization) \cite{smdd}. We use a non-agentic evaluation in which both the lead and generated candidate are scored with ADMET-AI and RDKit. We compare direct generation (\emph{base}) and MH refinement (\emph{MH}) using the original proposal ($q_0$) and the proposal after one distillation round ($q_1$), against the RL-trained baseline C-MORAL. Since C-MORAL's task objectives do not align with SMDD Task 4, we pool the five IND seed files and train one shared adapter per backbone. Each seed retains its own improvement and hold-constant keys, while the oracle scores the union of IND properties.

\paragraph{Metrics.}
Following C-MORAL~\citep{cmoral}, we report Success Optimized Rate (SOR), the proportion of molecules that meet the requested property-improvement criteria, and Strict Success Optimized Rate (SSOR), the proportion whose relevant properties all satisfy prescribed near-optimal thresholds. For SMDD-Bench~\citep{smdd}, an SOR success requires every requested objective to improve by its instance-specific threshold. Similarity (Sim) is the Morgan fingerprint~\citep{morgan1965,rogers2010ecfp,rdkit} Tanimoto similarity~\citep{bajusz2015tanimoto,maggiora2014similarity} between each generated molecule and its lead. Clearance is the rate at which the body, mainly the liver, removes a drug from plasma~\citep{yim2021clearance}. For the 144 SMDD-Bench instances requesting reduced clearance, CLR is the fraction whose selected molecule's predicted clearance decreases by at least the instance-specific threshold~\citep{swanson2024admetai}. We select the best of 20 candidates by SOR and emphasize SSOR because it captures simultaneous satisfaction of all property targets at near-optimality.
\subsection{Main Results}

\begin{table*}[t]
\centering
\small
\setlength{\tabcolsep}{4.5pt}
\caption{
C-MuMOInstruct results on 100 held-out leads per task.
\emph{$q_0$} denotes the original proposal and $q_1$ the proposal after one distillation round; \emph{base} denotes direct generation and \emph{MH} refinement.
Best-of-$20$ selection prioritizes SOR, then SSOR, then similarity.
Success--similarity tradeoffs are shown in Fig.~\ref{fig:ssor-coverage}.
Bold marks the best value in each row.
}
\label{tab:backbone-main}
\medskip
\begin{tabular}{lllccccc}
\toprule
Backbone & Metric & Split
& C-MORAL
& $q_0$-base
& $q_0$-MH
& $q_1$-base
& $q_1$-MH \\
\midrule

\multirow{4}{*}{GeLLMO-Llama}
& \multirow{2}{*}{SOR $\uparrow$}
& IND & 0.296 & 0.368 & 0.630 & 0.486 & \textbf{0.696} \\
& & OOD & 0.316 & 0.305 & 0.506 & 0.390 & \textbf{0.548} \\
\cmidrule(lr){2-8}
& \multirow{2}{*}{SSOR $\uparrow$}
& IND & 0.054 & 0.074 & 0.170 & 0.130 & \textbf{0.244} \\
& & OOD & 0.186 & 0.163 & 0.251 & 0.230 & \textbf{0.286} \\
\midrule

\multirow{4}{*}{GeLLMO-Mistral}
& \multirow{2}{*}{SOR $\uparrow$}
& IND & 0.530 & 0.354 & 0.590 & 0.472 & \textbf{0.706} \\
& & OOD & 0.390 & 0.271 & 0.467 & 0.332 & \textbf{0.552} \\
\cmidrule(lr){2-8}
& \multirow{2}{*}{SSOR $\uparrow$}
& IND & 0.110 & 0.068 & 0.126 & 0.124 & \textbf{0.210} \\
& & OOD & 0.222 & 0.160 & 0.241 & 0.212 & \textbf{0.298} \\
\midrule

\multirow{4}{*}{mCLM-4B}
& \multirow{2}{*}{SOR $\uparrow$}
& IND & 0.461 & 0.162 & 0.394 & 0.266 & \textbf{0.496} \\
& & OOD & 0.283 & 0.158 & 0.338 & 0.262 & \textbf{0.408} \\
\cmidrule(lr){2-8}
& \multirow{2}{*}{SSOR $\uparrow$}
& IND & 0.054 & 0.018 & 0.078 & 0.048 & \textbf{0.120} \\
& & OOD & 0.126 & 0.058 & 0.158 & 0.110 & \textbf{0.224} \\
\bottomrule
\end{tabular}
\end{table*}

\paragraph{Refinement across backbones.}
\label{sec:backbones}
Table~\ref{tab:backbone-main} shows effect of inference-time refinement across three model families on both IND/OOD splits. MH improves $q_0$ SOR by $\approx 80\%$ and SSOR by $\approx 90\%$ on average. The same pattern persists after distillation, with $q_1$-MH outperforming $q_1$-base on every backbone and split. These gains show that substantial optimization capability is already present in the proposal models but is not recovered by one-shot decoding; sequential MH refinement exposes this capability by composing and selecting edits under the target objective.

\begin{figure*}[t]
\centering
\includegraphics[width=\textwidth]{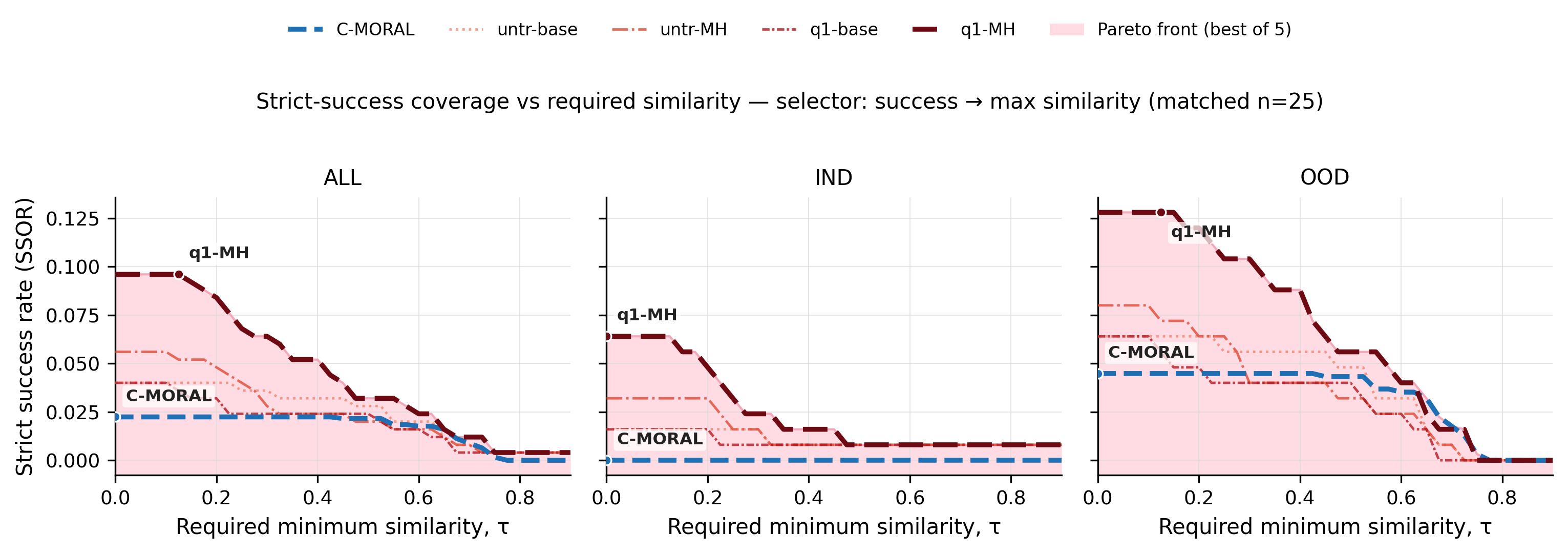}
\caption{
Strict-success coverage versus minimum lead similarity on 25 matched leads per task.
Each method selects candidates by SOR success, then maximum similarity; coverage counts selected outputs that also satisfy SSOR and the indicated similarity threshold.
C-MORAL OOD results are averaged over five task-specific GRPO policies.
}
\label{fig:ssor-coverage}
\end{figure*}

\paragraph{Success under similarity requirements.}
Figure~\ref{fig:ssor-coverage} evaluates the tradeoff between strict multi-objective success and similarity to the starting lead.
Across increasingly stringent similarity requirements, \modelname{} forms empirical Pareto frontier in SSOR--similarity space, maintaining higher strict-success coverage than C-MORAL over the evaluated range. This advantage holds on the pooled evaluation and extends to the held-out OOD property combinations. Because SSOR requires all relevant properties to reach their prescribed near-optimal thresholds~\citep{cmoral}, the frontier measures how often a method can satisfy all molecular objectives while remaining close to the original lead.
These results show that \modelname{} improves the strict-success--locality tradeoff rather than obtaining higher success simply through larger structural changes.

\paragraph{Search and proposal distillation.}
\label{sec:distillation-results}
Table~\ref{tab:backbone-main} shows that proposal distillation substantially strengthens the base model. On GeLLMO-Mistral, one distillation round improves zero-shot base SOR by \textbf{$33\%$} on IND ($0.354\!\rightarrow\!0.472$) and \textbf{$23\%$} on OOD ($0.271\!\rightarrow\!0.332$). The trained base model also improves subsequent MH refinement by \textbf{$20\%$} on IND ($0.590\!\rightarrow\!0.706$) and \textbf{$18\%$} on OOD ($0.467\!\rightarrow\!0.552$). The same pattern holds for GeLLMO-Llama, where distillation improves zero-shot base by \textbf{$32\%$/$28\%$} and MH refinement by \textbf{$10\%$/$8\%$} on IND/OOD. These results suggest not only could inference-time MH improve multi-objective optimization: distilling search trajectories back into the base CLM produces an improved base model for both zero-shot generation and subsequent search.

\begin{table*}[t]
\centering
\caption{SMDD-Bench Task 4 results on 314 instances. ReMedI-MH uses the frozen proposal, and pooled C-MORAL uses one shared adapter per backbone at the same 20-candidate inference budget.}
\label{tab:smdd-main}
\medskip
\begin{tabular}{lcccccc}
\toprule
Backbone & \multicolumn{2}{c}{SOR $\uparrow$}
& \multicolumn{2}{c}{Sim $\uparrow$}
& \multicolumn{2}{c}{CLR $\uparrow$} \\
& Ours & C-MORAL & Ours & C-MORAL & Ours & C-MORAL \\
\midrule
GeLLMO-Llama
& \textbf{0.764} & 0.580 & 0.633 & \textbf{0.707} & \textbf{0.764} & 0.583 \\
\midrule
GeLLMO-Mistral
& \textbf{0.771} & 0.640 & 0.663 & \textbf{0.737} & \textbf{0.792} & 0.681 \\
\midrule
mCLM-4B
& \textbf{0.796} & 0.506 & 0.316 & \textbf{0.361} & \textbf{0.806} & 0.576 \\
\bottomrule
\end{tabular}
\end{table*}

\paragraph{Generalization to SMDD-Bench.}
On 314 SMDD-Bench instances, ReMedI-MH improves SOR by $0.13$--$0.29$ and clearance-target success on every backbone compared with pooled C-MORAL, while C-MORAL retains higher similarity (Table~\ref{tab:smdd-main}). ReMedI's shared adapter uses trajectories from 500 lead--task instances, about $7.8\times$ fewer than C-MORAL's 3,883 RL training prompts~\citep{cmoral}.

\begin{figure*}[t]
    \centering
    \includegraphics[width=0.80\textwidth]{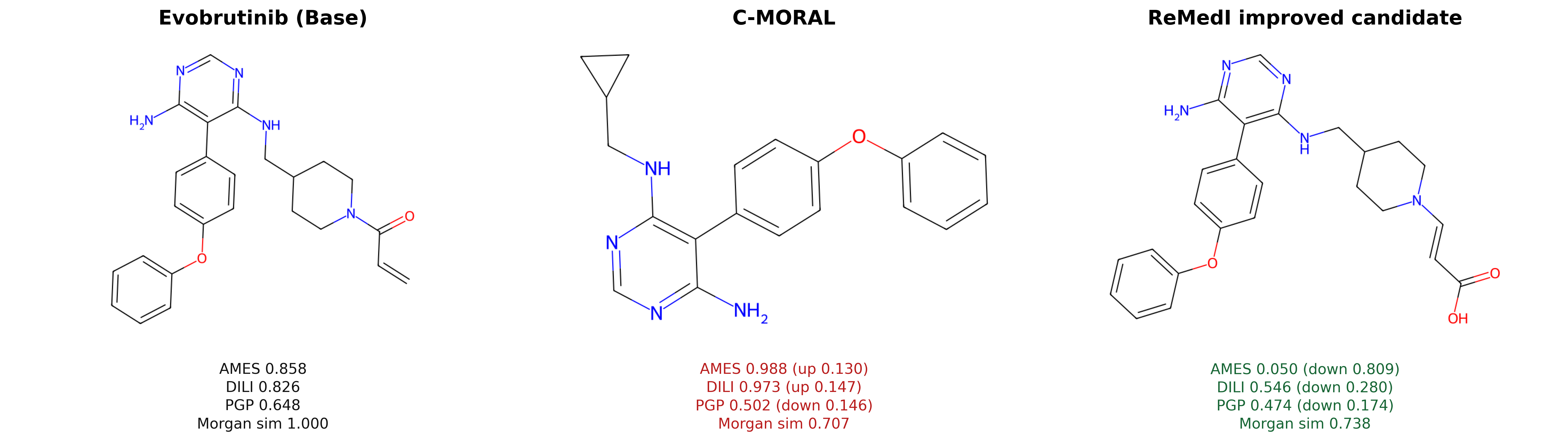}
    \caption{
    \textbf{Evobrutinib case study.}
    Starting from the same ``fallen angel'' drug lead~\cite{mclm}, ReMedI identifies a candidate that simultaneously reduces all three predicted liabilities while preserving substantial structural similarity. This example illustrates how inference-time refinement can incorporate multiple beneficial edits into a single molecule while retaining the lead scaffold.
    }
    \label{fig:evobrutinib}
\end{figure*}

\paragraph{Evobrutinib case study.}
Figure~\ref{fig:evobrutinib} compares optimization of the failed drug Evobrutinib. ReMedI simultaneously reduces AMES ($0.858\!\rightarrow\!0.050$), DILI ($0.826\!\rightarrow\!0.546$), and PGP ($0.648\!\rightarrow\!0.474$) while retaining similarity $0.738$. By contrast, C-MORAL improves PGP but worsens DILI and AMES, with similarity $0.707$. Whereas the mCLM example applies sequential property-specific edits, \modelname{} finds a single search-derived candidate that improves all three liabilities simultaneously.

\section{Ablations}
\label{sec:ablation}
\begin{wrapfigure}{r}{0.48\textwidth}
    \centering
    \includegraphics[width=\linewidth]{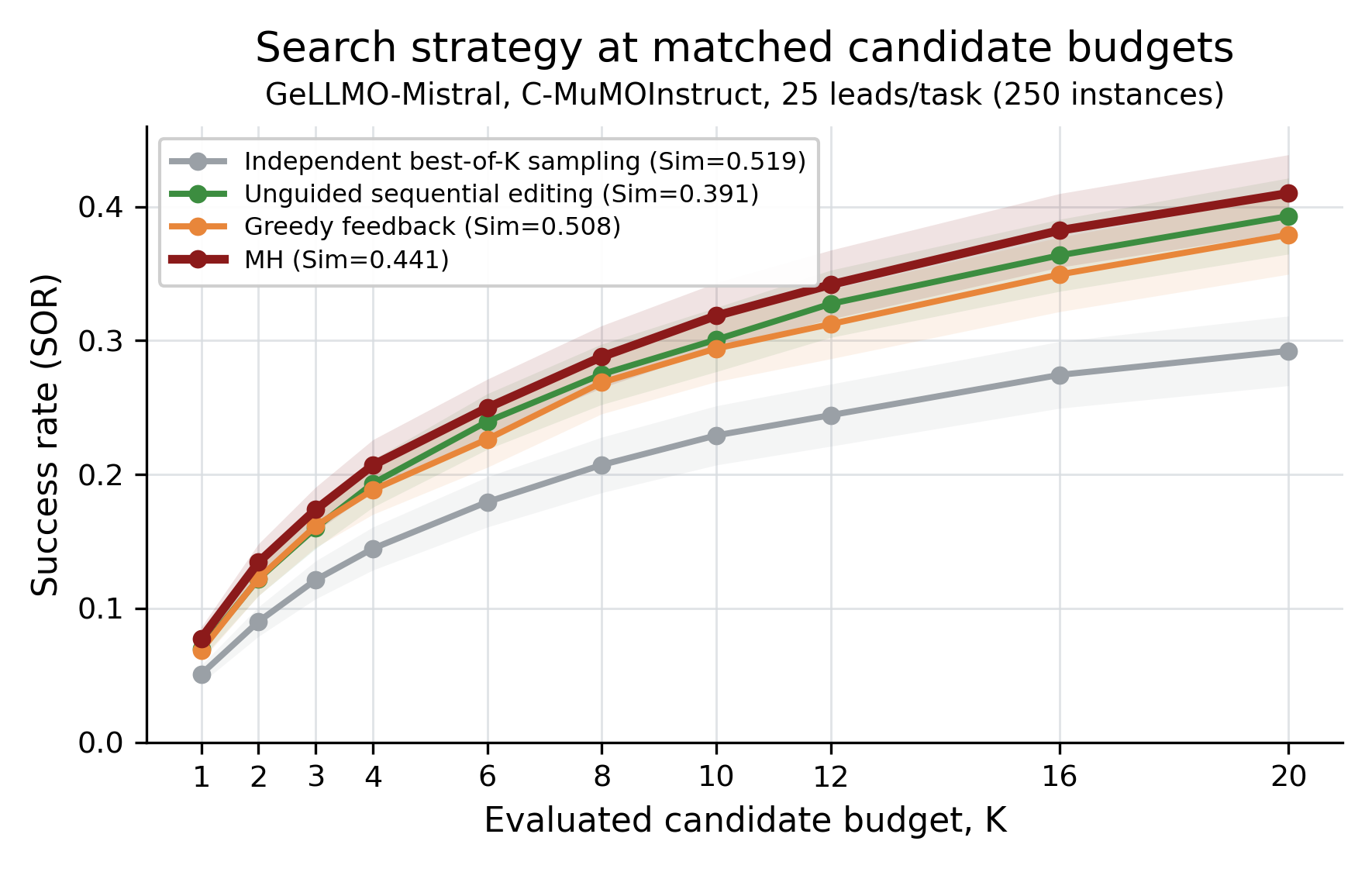}
    \caption{
    SOR versus candidate budget on 250 GeLLMO-Mistral instances spanning ten C-MuMOInstruct tasks.
    At matched candidate budgets, MH achieves the highest observed SOR.
    Legend values report mean lead similarity.
    }
    \vspace{-4pt}
    \label{fig:search-budget}
\end{wrapfigure}

\paragraph{Effect of search strategy.}
Figure~\ref{fig:search-budget} compares four search strategies under the same evaluated candidate budget:
(i) \emph{Independent best-of-$K$}, which samples every candidate directly from zero-shot generations;
(ii) \emph{unguided sequential editing}, which repeatedly conditions on the most recent molecule and always advances to the next proposal;
(iii) \emph{greedy feedback}, which edits sequentially but always accepts proposals that improve the objective; and
(iv) \emph{MH}, which is our method. Sequential methods all outperform independent sampling, among those, MH achieves the highest SOR at every tested budget, reaching 0.41 at $K=20$ compared with 0.39 for unguided editing and 0.38 for greedy feedback. MH also preserves greater lead similarity than unguided editing (0.44 vs. 0.39), indicating that its gain does not come simply from allowing larger structural changes. Overall, the figure demonstrates the advantage of sequential edits over one-shot sampling, and that MH gives the best success--similarity tradeoff among the sequential strategies.

\begin{figure}[t]
    \centering
    \begin{minipage}[b]{0.42\linewidth}
        \centering
        \includegraphics[width=\linewidth]{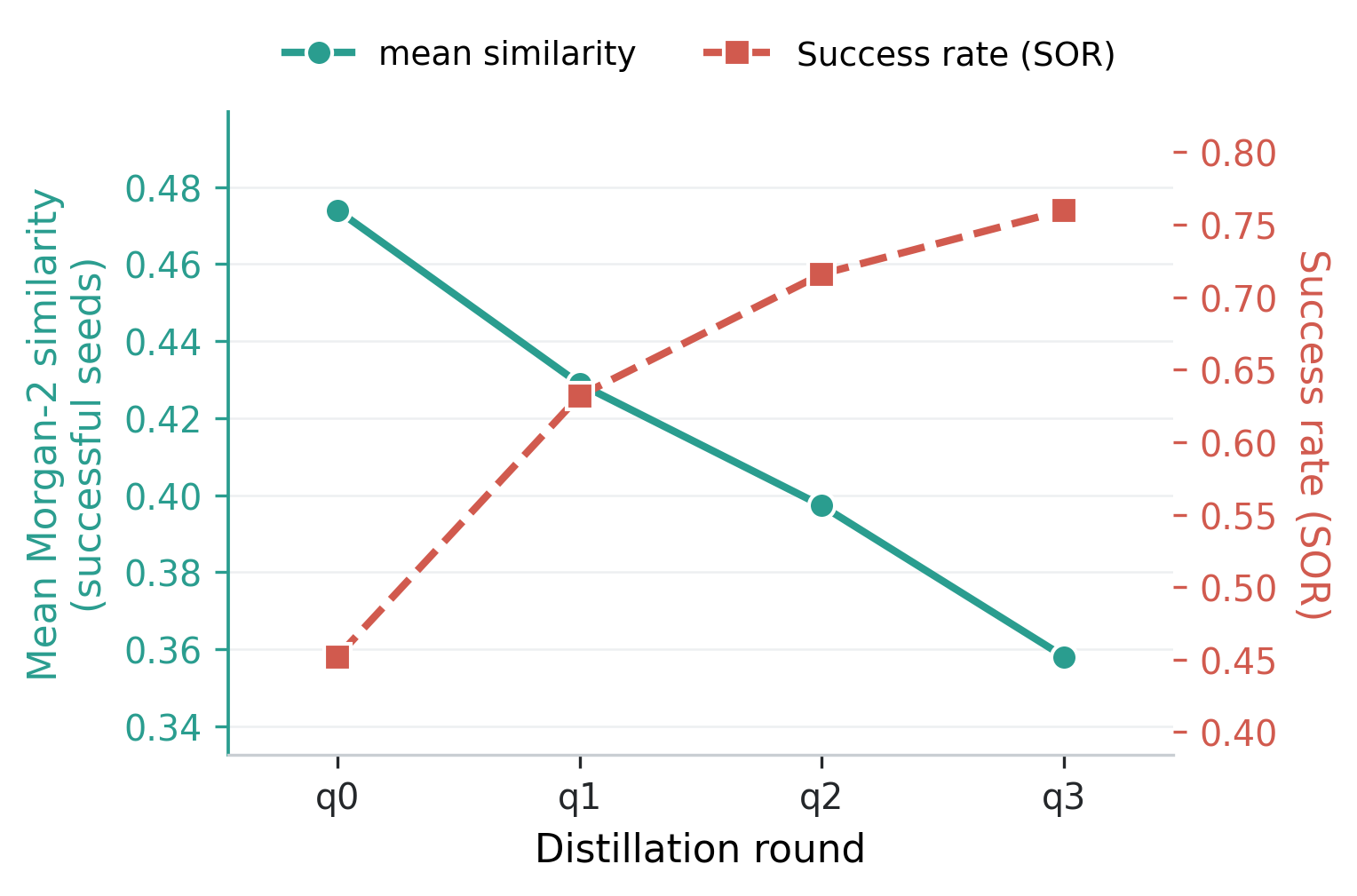}
        \par\smallskip
        {\small (a) SOR and similarity.}
    \end{minipage}
    \hspace{0.04\linewidth}
    \begin{minipage}[b]{0.42\linewidth}
        \centering
        \includegraphics[width=\linewidth]{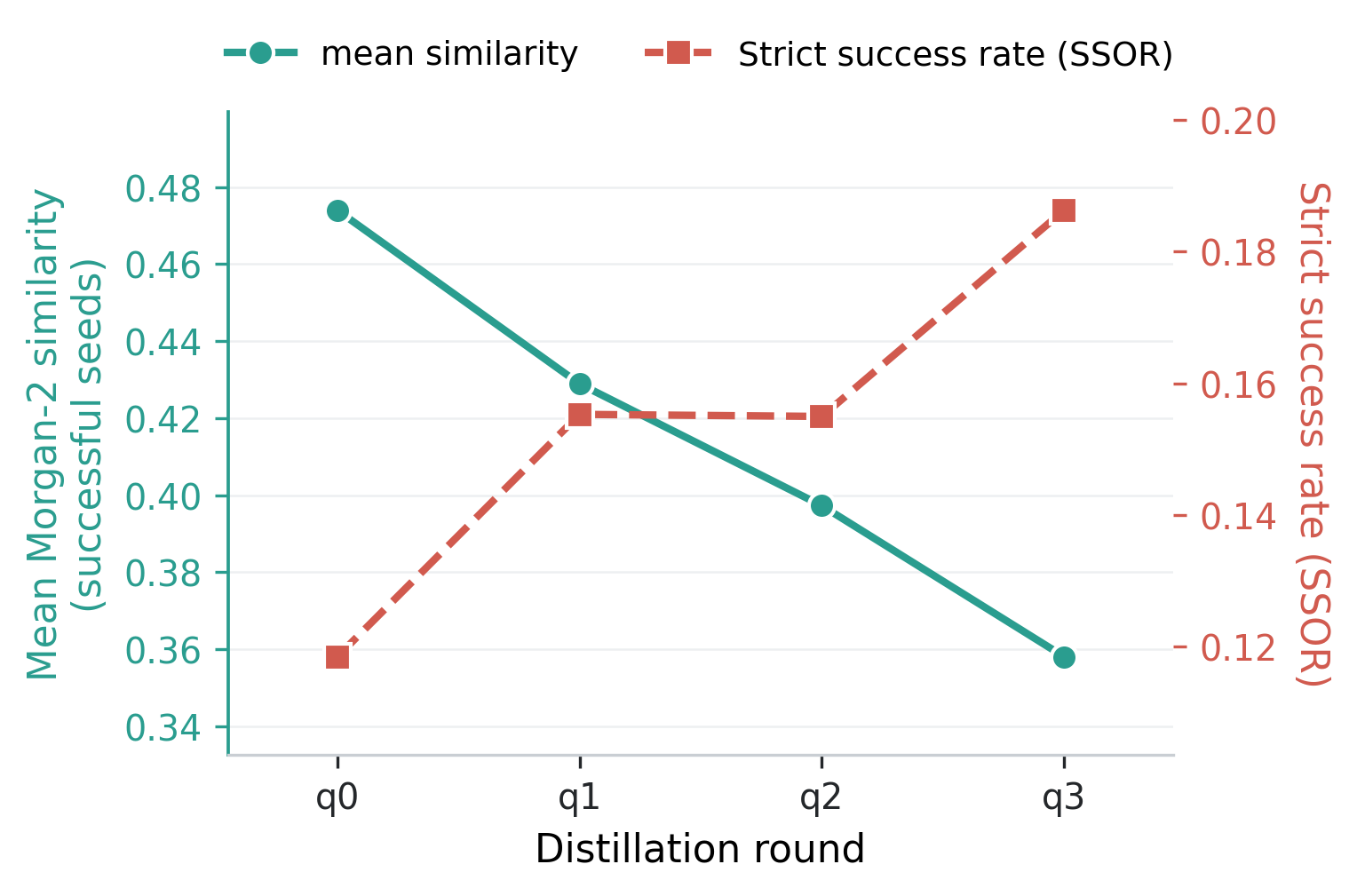}
        \par\smallskip
        {\small (b) SSOR and similarity.}
    \end{minipage}
    \caption{Effect of repeated proposal distillation. Red dashed lines show SOR/SSOR; blue lines show mean lead similarity. $q_0$ is the original proposal and $q_1$--$q_3$ are successive distillation rounds.
    }
    \label{fig:proposal-iteration}
\end{figure}

\paragraph{Choosing the distillation round.}
Figure~\ref{fig:proposal-iteration} shows that the first distillation round captures the largest useful gain from search-derived supervision. From $q_0$ to $q_1$, SOR increases from approximately $0.45$ to $0.63$ and SSOR from $0.12$ to $0.16$, with only a modest decrease in similarity from $0.47$ to $0.43$. Additional rounds provide diminishing returns: $q_2$ improves SOR but not SSOR, while $q_3$ raises SSOR further only as similarity falls to approximately $0.36$.
We therefore use \textbf{$q_1$} as our primary proposal, as it captures the largest simultaneous improvement in SOR and SSOR before later distillation rounds increasingly trade locality for additional optimization.

\begin{figure}[t]
    \centering
    \includegraphics[width=0.80\linewidth]{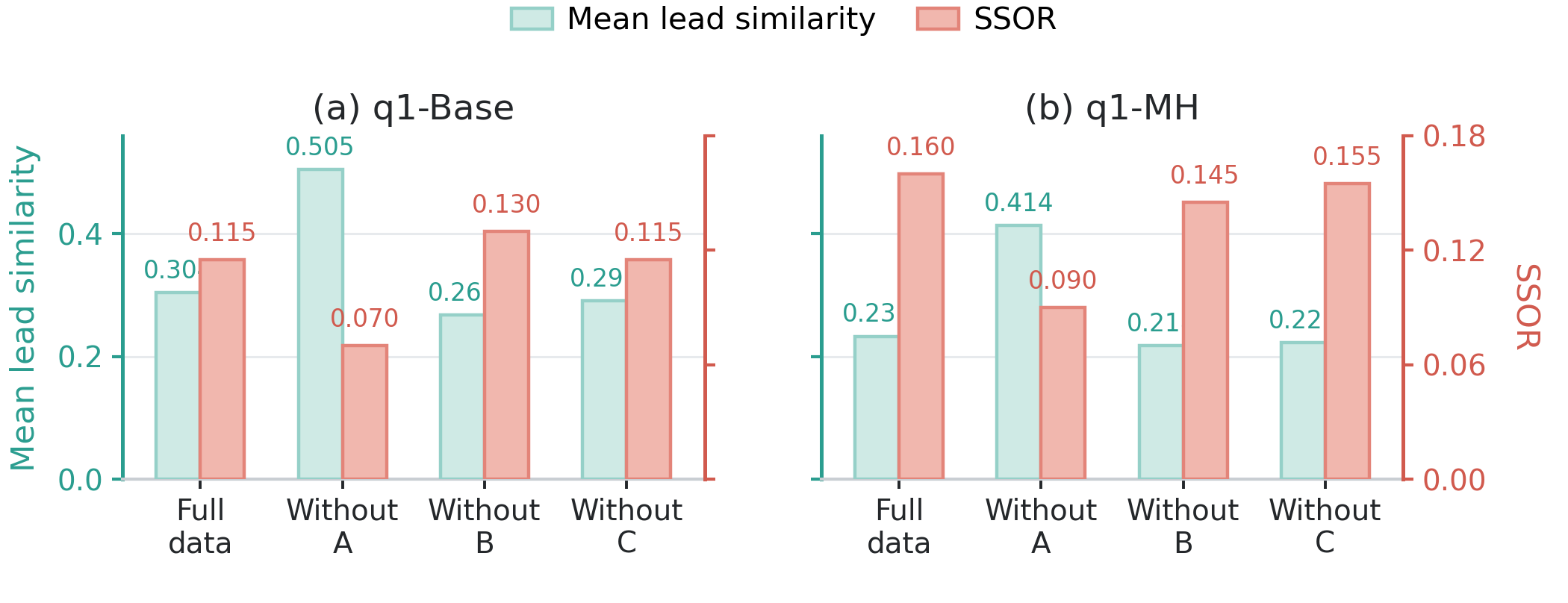}
    \caption{Effect of distillation sources on $q_1$-Base and $q_1$-MH. $-$A, $-$B, and $-$C remove successful endpoints, intermediate states, and improving transitions, respectively. Blue shows mean similarity; red dashed shows SSOR.
    }
    \label{fig:distillation-sources}
\end{figure}

\paragraph{Effect of distillation sources.}
Figure~\ref{fig:distillation-sources} ablates the three supervision sources used for proposal distillation:
(A) \emph{lead-conditioned successful endpoints}, which teach complete solutions from the original lead;
(B) \emph{intermediate-state examples}, which train the proposal on states encountered during refinement; and
(C) \emph{accepted transitions}, which teach useful local edits that do not yet satisfy SOR.
Removing (A) causes the largest drop in SSOR, showing that endpoint supervision is essential for learning successful optimization. Ablation (A) makes the proposal much more conservative: for q1-Base, similarity rises from $0.30$ to $0.505$ while SSOR drops from $0.115$ to $0.070$; after MH refinement, similarity rises from $0.23$ to $0.414$ while SSOR falls from $0.160$ to $0.090$. Thus, lead-conditioned successful endpoints appear to provide the main supervision for learning edits that are large enough to satisfy the optimization targets. Removing (B) or (C) lowers refined SSOR while also reducing similarity. The full mixture gives the highest refined SSOR ($0.160$), supporting trajectory-level distillation and providing the strongest operating point for subsequent MH refinement. 

\section{Limitations \& Conclusion}
\label{sec:limits}
\modelname{} reframes molecular optimization as inference-time refinement: a CLM proposes sequential edits, MH selects them under an explicit lead-relative objective, and successful search trajectories are distilled back into the proposal. Across C-MuMOInstruct and SMDD-Bench, this improves frozen proposals and yields a stronger strict-success--similarity tradeoff, while one round of distillation further improves both direct generation and subsequent search. The ablations further support the role of each component: sequential editing provides the main search gain, MH improves the success--similarity tradeoff, and the three distillation sources contribute complementary supervision for direct generation and subsequent refinement. A limitation of our method is all discoveries remains limited \textit{in silico} predictions, which do not yet establish \textit{in vitro} or \textit{in vivo} efficacy. 
\label{sec:conclusion}

\subsection*{AI use statement}
Generative AI tools assisted with prose editing and routine code and analysis. The authors are responsible for verifying claims, figures, and references before submission.

\bibliography{iclr2027_conference}
\bibliographystyle{iclr2027_conference}

\newpage

\appendix

\section{Appendix}
\subsection{Benchmark Metrics}
Let $f_r(x)$ denote property $r$ of molecule $x$, and let $s_r\in\{+1,-1\}$ indicate whether higher or lower values are preferred. For lead $x_{0,i}$ and output $x_i$, SOR requires each targeted property to improve by its prescribed margin $\Delta_r$ and each protected property to remain within tolerance $\epsilon_r$:
\begin{equation}
\mathrm{SOR}
=
\frac{1}{n}\sum_{i=1}^{n}
\prod_{r\in\mathcal{P}_{\mathrm{improve}}}
\mathbf{1}\{s_r(f_r(x_i)-f_r(x_{0,i}))\geq\Delta_r\}
\prod_{r\in\mathcal{P}_{\mathrm{preserve}}}
\mathbf{1}\{|f_r(x_i)-f_r(x_{0,i})|\leq\epsilon_r\}.
\label{eq:sor}
\end{equation}
SSOR instead checks whether all relevant properties meet their prescribed near-optimal thresholds $\Theta_r$~\citep{cmoral}:
\begin{equation}
\mathrm{SSOR}
=
\frac{1}{n}\sum_{i=1}^{n}
\prod_{r\in\mathcal{P}}
\mathbf{1}\!\left[
s_r\bigl(f_r(x_i)-\Theta_r\bigr)\geq0
\right].
\end{equation}

For example, consider hypothetical normalized permeability and toxicity scores, with targets of permeability $\geq0.8$ and toxicity $\leq0.2$.
Suppose the lead has permeability $0.4$ and toxicity $0.1$, and the request requires a permeability increase of at least $0.1$ while preserving toxicity.
An output with permeability $0.6$ and unchanged toxicity passes SOR but fails SSOR: permeability improves sufficiently but remains below its absolute target.
An output with permeability $0.85$ and unchanged toxicity passes both. 


\begin{table*}[t]
\centering
\small
\setlength{\tabcolsep}{4.5pt}
\caption{C-MuMOInstruct performance of C-MORAL and \modelname{} across GeLLMO-C (Llama, Mistral) and mCLM-4B backbones. $n=50$ held-out seeds/task, best-of-$20$. \emph{$q_0$} is the frozen proposal, \emph{$q_1$} one round of search distillation, \emph{base} direct generation, and \emph{MH} inference-time refinement. This 50-lead-per-task summary uses the candidate selection in the source draft (success, then strict success, then similarity), distinct from the success-then-maximum-similarity selector of the 25-lead coverage figures. SOR, SSOR, and Sim are reported across methods; RI is a within-method diagnostic and is not comparable across methods (C-MORAL's RI is computed on a different scale), so RI rows are not bolded. Bold marks the best comparable value in each row, excluding \emph{$q_0$-base}.}
\medskip{}
\label{tab:backbone_full}
\begin{tabular}{lllccccc}
\toprule
Backbone & Metric & Split
& C-MORAL
& $q_0$-base
& $q_0$-MH
& $q_1$-base
& $q_1$-MH \\
\midrule
\multirow{8}{*}{GeLLMO-Llama}
& \multirow{2}{*}{SOR $\uparrow$}
& IND & 0.320 & 0.340 & 0.632 & 0.532 & \textbf{0.756} \\
& & OOD & 0.290 & 0.300 & 0.448 & 0.424 & \textbf{0.524} \\
\cmidrule(lr){2-8}
& \multirow{2}{*}{SSOR $\uparrow$}
& IND & 0.030 & 0.092 & 0.200 & 0.140 & \textbf{0.264} \\
& & OOD & 0.140 & 0.180 & 0.276 & 0.240 & \textbf{0.332} \\
\cmidrule(lr){2-8}
& \multirow{2}{*}{RI (diag.)}
& IND & 0.61 & 0.23 & 0.64 & 0.49 & 1.14 \\
& & OOD & 0.52 & 0.31 & 0.51 & 0.49 & 0.67 \\
\cmidrule(lr){2-8}
& \multirow{2}{*}{Sim $\uparrow$}
& IND & 0.460 & 0.686 & 0.491 & \textbf{0.523} & 0.369 \\
& & OOD & 0.450 & 0.686 & \textbf{0.596} & 0.569 & 0.487 \\
\midrule

\multirow{8}{*}{GeLLMO-Mistral}
& \multirow{2}{*}{SOR $\uparrow$}
& IND & 0.400 & 0.324 & 0.528 & 0.524 & \textbf{0.692} \\
& & OOD & 0.322 & 0.268 & 0.400 & 0.372 & \textbf{0.532} \\
\cmidrule(lr){2-8}
& \multirow{2}{*}{SSOR $\uparrow$}
& IND & 0.056 & 0.068 & 0.124 & 0.120 & \textbf{0.196} \\
& & OOD & 0.122 & 0.164 & 0.236 & 0.280 & \textbf{0.356} \\
\cmidrule(lr){2-8}
& \multirow{2}{*}{RI (diag.)}
& IND & 4.85 & 0.28 & 0.43 & 0.41 & 0.72 \\
& & OOD & 4.61 & 0.19 & 0.38 & 0.33 & 0.64 \\
\cmidrule(lr){2-8}
& \multirow{2}{*}{Sim $\uparrow$}
& IND & 0.565 & 0.719 & \textbf{0.595} & 0.586 & 0.467 \\
& & OOD & 0.582 & 0.722 & \textbf{0.644} & \textbf{0.644} & 0.533 \\
\midrule

\multirow{8}{*}{mCLM-4B}
& \multirow{2}{*}{SOR $\uparrow$}
& IND & 0.420 & 0.168 & 0.432 & 0.220 & \textbf{0.448} \\
& & OOD & 0.273 & 0.152 & \textbf{0.356} & 0.212 & \textbf{0.356} \\
\cmidrule(lr){2-8}
& \multirow{2}{*}{SSOR $\uparrow$}
& IND & 0.044 & 0.012 & 0.088 & 0.068 & \textbf{0.156} \\
& & OOD & 0.116 & 0.080 & 0.204 & 0.096 & \textbf{0.216} \\
\cmidrule(lr){2-8}
& \multirow{2}{*}{RI (diag.)}
& IND & 5.78 & 0.06 & 0.39 & 0.13 & 0.47 \\
& & OOD & 6.85 & 0.13 & 0.43 & 0.19 & 0.44 \\
\cmidrule(lr){2-8}
& \multirow{2}{*}{Sim $\uparrow$}
& IND & 0.235 & 0.421 & 0.363 & \textbf{0.406} & 0.363 \\
& & OOD & 0.231 & 0.388 & 0.360 & \textbf{0.409} & 0.384 \\
\bottomrule
\end{tabular}
\end{table*}

\begin{table}[t]
\centering
\caption{
Effect of trajectory depth on frozen GeLLMO-Mistral.
Every configuration uses 80 proposals per lead.
Results aggregate IND and OOD; bold marks the best value in each column.
}
\medskip
\label{tab:depth-summary}
\begin{tabular}{lccc}
\toprule
Chains$\times$steps & SOR $\uparrow$ & SSOR $\uparrow$ & Sim $\uparrow$ \\
\midrule
$80\times1$ & 0.440 & 0.070 & \textbf{0.463} \\
$40\times2$ & 0.500 & 0.100 & 0.422 \\
$20\times4$ & \textbf{0.560} & 0.125 & 0.372 \\
$10\times8$ & 0.550 & \textbf{0.140} & 0.343 \\
\bottomrule
\end{tabular}
\end{table}

\subsection{Performance across backbones.}
Table~\ref{tab:backbone_full} reports the full C-MuMOInstruct results across GeLLMO-Llama, GeLLMO-Mistral, and mCLM-4B. Inference-time refinement consistently improves the frozen proposal: $q_1$-MH increases both SOR and SSOR over $q_1$-base for every backbone and split. Distillation further strengthens the proposal on the GeLLMO backbones, and combining one round of distillation with refinement ($q_1$-MH) gives the highest SOR and SSOR in nearly all settings. The gains also extend to OOD property combinations, indicating that trajectories distilled from IND tasks improve optimization beyond the combinations used for training. Similarity generally decreases as optimization becomes more aggressive, motivating the similarity-constrained coverage analysis in the main text rather than interpreting success and similarity independently. RI is included only as a within-method diagnostic because its scale is not comparable to the C-MORAL implementation.

\subsection{Trajectory Depth Ablation}
\label{sec:depth}
Table~\ref{tab:depth-summary} compares trajectory depths at a fixed budget of 80 proposals per lead.
Independent proposals preserve locality but rarely reach the stringent target.
Replacing 80 independent proposals with 10 trajectories of eight steps doubles overall SSOR from $0.070$ to $0.140$.
Because every configuration evaluates the same number of proposals, this gain cannot be attributed to additional sampling.
Instead, sequential conditioning allows multiple edits to compose into solutions satisfying the stringent multi-property constraints.

\end{document}

%% file: math_commands.tex
\usepackage{amsmath,amsfonts,bm}

\def\eqref#1{equation~\ref{#1}}

\def\1{\bm{1}}

\DeclareMathAlphabet{\mathsfit}{\encodingdefault}{\sfdefault}{m}{sl}
\SetMathAlphabet{\mathsfit}{bold}{\encodingdefault}{\sfdefault}{bx}{n}

\DeclareMathOperator*{\argmax}{arg\,max}
\DeclareMathOperator*{\argmin}{arg\,min}